\documentclass[10pt, a4paper]{article}
\pdfoutput=1 
\usepackage{lrec}
\usepackage{multibib}
\newcites{languageresource}{Language Resources}
\usepackage{graphicx}
\usepackage{tabularx}
\usepackage{soul}

\usepackage{epstopdf}
\usepackage[T1]{fontenc}
\usepackage[utf8]{inputenc}

\usepackage{relsize}
\usepackage{booktabs}

\usepackage{hyperref}
\usepackage{xstring}

\usepackage{pgfplots}

\newcommand{\conllu}[0]{{CoNLL-U}}

\title{NoReC: The Norwegian Review Corpus}

\name{Erik Velldal, Lilja Øvrelid, Eivind Alexander Bergem, Cathrine Stadsnes,  \\ 
{\bf \large Samia Touileb, Fredrik Jørgensen$^\dagger$}}

\address{Language Technology Group, Department of Informatics, University of Oslo\\
  \{erikve, liljao, eivinabe, cathsta, samiat\}@ifi.uio.no\\
  $^\dagger$Schibsted Media Group\\
  fredrik.jorgensen@schibsted.com\\} 

\abstract{This paper presents the Norwegian Review Corpus (NoReC), created for
  training and evaluating models for document-level sentiment analysis. The
  full-text reviews have been collected from major Norwegian news sources and
  cover a range of different domains, including literature, movies, video
  games, restaurants, music and theater, in addition to product reviews across
  a range of categories.  Each review is labeled with a manually assigned score
  of 1--6, as provided by the rating of the original author.  This first
  release of the corpus comprises more than 35,000 reviews. 
  It is distributed using the \conllu\ format, pre-processed using
  UDPipe, 
  along with a rich set of metadata. 
  The work reported in this paper forms part of the SANT initiative (Sentiment
  Analysis for Norwegian Text), a project seeking to provide resources and
  tools for sentiment analysis and opinion mining for Norwegian.  As resources
  for sentiment analysis have so far been unavailable for Norwegian, NoReC
  represents a highly valuable and sought-after addition to Norwegian language
  technology. \\ \newline }

\begin{document}

\maketitleabstract

\section{Introduction} 

Norwegian is in many ways an under-resourced language, with training
and evaluation data still lacking for many core NLP tasks. The current
work aims to alleviate the situation for the particular task of
sentiment analysis.  The SANT project -- Sentiment Analysis for
Norwegian Text -- seeks to create, and make publicly available,
resources and tools for sentiment analysis for Norwegian. The SANT
effort described in the current paper marks the release of the
Norwegian Review Corpus\footnote{\url{https://github.com/ltgoslo/norec}} 
(NoReC). The dataset comprises more than 
35,000 full-text reviews from a range of different domains, collected
from several of the major Norwegian news sources. Each review is rated
with a numerical score on a scale of 1--6, and can be used for
training and evaluating models for document-level sentiment analysis.

\subsection{Rating by dice} 

A particularity of review journalism in Norway, is the wholesale adoption of
dice rolls (`terningkast') as a standard rating scale: The item under review is
rated on a scale of 1--6, commonly visualized by the face of a % white-on-red
die with a corresponding number of `dots' or pips. 
The practice is thought to have been 
introduced 
already in 1952 when reviewing movies in the  
newspaper \emph{Verdens Gang (VG)}. By now it is has found
widespread use in all sorts of arts and consumer journalism and is used when reviewing
everything from books, theater and music, to home electronics,
restaurants, and children's clothing.   

The rating practice described above has several benefits for the goal of 
document-level SA: (i) It eliminates the need for costly manual annotation
since the numerical rating (i.e., the die roll) directly provides us with the
labels needed for training models for detecting the overall document
polarity. (ii) There is no need for manually defined mappings to 
align different rating schemes as the reviews all use a uniform scale. (iii) The
wide range of available news sources using the same rating practice, including
all the major national newspapers, facilitates the creation of a large-scale
dataset. 
(iv) Models trained on the dataset can be expected to generalize well across
domains given the balance of different topics covered in the corpus.  For
English, a substantial amount of SA research has been directed towards
predicting the sentiment of movie reviews, collected from aggregator sites like
IMDb.com and RottenTomatoes.com \cite{Pan:Lee:05,Soc:Per:Wu:13} or other types of consumer reviews \cite{Bli:Dre:Per:07,Wan:Lu:Zha:10,Maa:Dal:Pha:11}. The NoRec data set contains a variety of different types of reviews from a range of domains.

\subsection{Sources and partners}\label{sec:sources}

The SANT project represents a newly initiated collaboration between the
Language Technology Group (LTG) at the Department of Informatics at the
University of Oslo, and three of Norway's largest media groups; the Norwegian
Broadcasting Corporation (NRK -- the state-owned public broadcaster) and the
privately held Schibsted Media Group and Aller Media.  This first release of
NoReC comprises 35,194 
reviews extracted from eight different news sources as contributed by the three
media partners.  In terms of publishing date the reviews mainly cover the time
span 2003--2017, although it also includes a handful of reviews dating back as
far as 1998.  We briefly present the sources provided by the different
partners below.

\paragraph{Schibsted Media Group} The Schibsted group has contributed content from their 
full portfolio of Norwegian news sources: VG, Aftenposten, Fædrelandsvennen,
Bergens Tidende, and Stavanger Aftenblad.  While the latter three rank among
Norway's largest regional newspapers, Aftenposten is the largest national
newspaper in terms of circulation and VG is the largest online news source with
more than 2.4 million readers across all platforms.

\paragraph{Aller Media} The Aller publishing company has contributed content
from two sources.  The first is the online version of the newspaper
Dagbladet -- the second most visited online news source in Norway -- publishing
reviews for music recordings and live performances, theater and related stage
performances, movies, literature, restaurants and more. 
The second source, DinSide.no, is a website specializing in product reviews,
covering a wide range of product types, from home electronics to cars and
clothing.

\paragraph{NRK} The Norwegian Broadcasting Corporation is a state-owned media
house, with a special mandate to be a non-commercial, politically independent
public broadcaster. For the review corpus, NRK has contributed content from the
website \href{http://p3.no/}{P3.no} which has an extensive back catalog 
of `die-rated' reviews of movies, TV series, computer games, and music (both
recordings and live performances).

Figure~\ref{tab:sources} shows the number of reviews included in the final
corpus, broken down across the various sources. 
\begin{table}
\centering
\begin{smaller}
\begin{tabular}{@{}lr@{}}
\toprule
Source & \# Reviews \\
\midrule
VG & 11,888 \\
Dagbladet & 5,305 \\
Stavanger Aftenblad & 5,146 \\
P3.no & 5,017 \\
DinSide.no & 2,944 \\
Fædrelandsvennen & 2,296 \\
Bergens Tidene & 1,675 \\
Aftenposten & 923 \\
\midrule
Total & 35,194 \\
\bottomrule
\end{tabular}
\caption{Number of reviews across sources.}
\label{tab:sources}
\end{smaller}
\end{table}

\section{Corpus creation}
\label{sec:corpus} 
 
The original document collections were provided from the media sources in
various JSON, HTML and XML formats, and a substantial effort has gone into
identifying relevant documents and extracting text and associated metadata. 

The extraction process can be summarized by the following three steps: (i)
Identify reviews, 
(ii) convert review content to an intermediate and canonical HTML format,  
(iii) extract text and pass it through linguistic pre-processing, producing 
representations in \conllu\ format, and finally (iv) extract relevant metadata to a JSON 
representation with normalized attribute--value names.  
We briefly comment on each of these steps in turn below. 

\subsection{Identifying reviews}
\label{sec:identification}
Some of the initial document dumps also included other articles beyond reviews,
and in these cases reviews had to be identified. While in some cases this could
be done simply by checking for an appropriate metadata field indicating the
rating score, other cases required checking for links pointing to an image of a
die (indicating the rating), or similar heuristics. 

For some of the sources, one and the same document may contain
multiple reviews, for example for product comparisons. In these cases 
we had to identify and separate out the different sub-reviews. Different
publishing conventions required targeting different types of cues 
in the document structure, like headers, bold-faced content or
die-face images. This also involved extraction of titles and rating
scores for the different sub-reviews. The identified sub-reviews
become separate documents in the NoReC data set. 

In total, 35,194 distinct reviews were extracted from the data provided by the
media partners.

\subsection{Converting content to canonical HTML}
\label{sec:content-to-html}

The raw data dumps from the sources are mostly in HTML format, but may
also be e.g. JSON objects, and have different conventions for document
structuring and use of mark-up.    
In order to streamline the downstream text extraction, all documents were converted to a
`canonical' HTML format where all textual content is located either inside a header
or a paragraph tag. 
In addition to containing the review text, the raw representations also
contains images, ads and other content not considered part of the running text.
In order to identify and mark the non-relevant text we used a combination of
heuristics based on simple string matching and properties like paragraph length
and ratio of content to markup.   
For example, care was taken to identify `you-might-also-be-interested-in' type
links that are injected throughout the texts in an attempt to keep the reader
on the website and generate more clicks.  Importantly, however, we chose not to
remove content when converting to our intermediate HTML format, instead 
introducing a new tag -- \texttt{remove} -- in which we enclose content
considered non-relevant. This non-destructive approach preserves the original
content, as to not close the door on changes to the subsequent task of text
extraction later.

\subsection{Linguistic enrichments and \conllu}
\label{sec:enrichments}

Given the canonical HTML format described above, it is straightforward to
extract the relevant text.  
In order to enable various types of downstream uses of the dataset, we further
preprocess the raw text using the UDPipe toolkit \cite{Str:Haj:Str:16},
representing each review as a \conllu\ file, following the format defined in
Universal Dependencies version
2.\footnote{\url{http://universaldependencies.org/format.html}} In this step we
perform sentence segmentation, tokenization, lemmatization, morphological
analysis, part-of-speech tagging and dependency parsing, following the
Universal Dependencies scheme \cite{Niv:Mar:Gin:16}.  However, the
preprocessing set-up is slightly complicated by the fact that the Norwegian
language has two official written standards -- Bokmål (the main variety) and
Nynorsk -- both of which are represented in the review corpus. 
Below we first describe how language identification is performed, and then go
on to give more details about UDPipe and the resulting \conllu\ data. 

\paragraph{Identifying language varieties}

The two official varieties are closely related and they are mostly
distinguished by minor lexical differences. Still, the differences are
strong enough that different preprocessing pipelines must be used for
the different standards, hence it is important to identify the
standard within a particular document.  Therefore, we have used
\texttt{langid.py} \cite{Lui:Bal:12} 
to identify the standard for each review.\footnote{\texttt{langid.py} can
  actually identify three different variants: \texttt{no}, \texttt{nn} and
  \texttt{nb}, for Norwegian (mixed), Nynorsk and Bokmål, respectively. While
  the precise details of how the classifier was trained are not clear, it
  appears to us after some experimentation that the classfication of Bokmål is
  more accurate when specyfing \texttt{no} rather than \texttt{nb} and hence is
  what we use here (together with \texttt{nn}). 
  We still use the language codes \emph{nb} and \emph{nn} when adding
  information about the detected standards to the metadata in NoReC.} 
We performed an evaluation of \texttt{langid.py} on 1599 reviews of which 1487 
were written in Bokmål 
and 112 in Nynorsk (based on selecting reviews from authors known to write in
a given variety). On this sample \texttt{langid.py} achieved 100\% accuracy. 
While the main variety, i.e. Bokmål, dominates the distribution in the corpus 
with 34,661 documents, we also identified 533 documents in Nynorsk 
(mainly from the sources Fædrelandsvennen, Bergens Tidende and P3.no). 

\paragraph{UDpipe configuration} 

We applied UDPipe \cite{Str:Haj:Str:16} v.1.2 with its pre-trained models for
Norwegian Bokmål and Nynorsk. This version of the UDPipe software
and the pre-trained models were developed for the CoNLL~2017 shared task 
\cite{Zem:Pop:Str:17}, which was devoted to parsing from raw text to
Universal Dependencies for more than 40 different languages. We use of the
models trained for participation in the shared task \cite{Str:Str:17}, 
not the models provided as baseline models for the participants.  
The Norwegian models were trained on the UD 2.0 versions of the
Norwegian UD treebanks \cite{Ovr:Hoh:16,Vel:Ovr:Hoh:17} in conjunction
with the aforementioned shared task, and the subsequent choice of
model for use (Bokmål vs Nynorsk) was determined by the language
identified for each particular review.

UDPipe obtained competitive results for Norwegian in the shared task, 
with rankings ranging between first place (lemmatization; both variants) and ninth place
(Bokmål dependency parsing LAS) out of 33 participating teams. In terms of
performance for the different sub-tasks, UDPipe reported 
F1 scores -- for Bokmål / Nynorsk respectively -- on sentence segmentation of
96.38 / 92.08, tokenization of 99.79 / 99.93, lemmatization of 96.66 / 96.48,
morphological analysis of 95.56 / 95.25, part-of-speech tagging of 96.83 /
96.54, and Labeled Accuracy Scores for dependency parsing of 83.89 / 82.74.

\paragraph{\conllu\ files}

\begin{table}
\centering
\begin{smaller}
\begin{tabular}{@{}lr@{}}
\toprule
 & \# \\
\midrule
Documents & 35,194 \\
Sentences & 837,914 \\
Tokens & 14,819,248 \\
Types, full-forms &  521,563 \\
Types, lemmas &  446,532 \\
\bottomrule
\end{tabular}
\caption{Basic corpus counts.}
\label{tab:corpus}
\end{smaller}
\end{table}

When extracting the text from the canonical HTML to pass it to UDPipe, we strip
away all mark-up and discard all content marked for removal as described in  
Section~\ref{sec:content-to-html} Double newlines were inserted between
paragraphs and excess whitespace trimmed away. Importantly, however, the text
structure is retained in \conllu\ by taking advantage of the support for
comments to mark paragraphs and sentences. In addition to the global document
ID number, each paragraph and sentence is also assigned a running ID within the
document, using the following form: 
\begin{itemize}
  \setlength{\parskip}{0pt}
  \setlength{\itemsep}{0pt plus 1pt}
  \item Paragraphs: \texttt{<review-id>-<paragraph-id>}, e.g. \texttt{000001-03} for paragraph 3 in document 1.
  \item Sentences: \texttt{<review-id>-<paragraph-id>-
  <sentence-id>}, e.g. \texttt{000001-03-02} for sentence 2 in paragraph 3 in document 1.
\end{itemize} 
After completing the UDPipe pre-processing, 
the corpus comprises a total of 
837,914 
sentences and 14,819,248 
tokens; see Table~\ref{tab:corpus} for an overview of some core corpus counts. 
A script for executing the entire pipeline from text extraction through UDPipe
parsing will be made available from the NoReC git repository. 

\subsection{Metadata and thematic categories}
\label{sec:metadata}

For all the identified reviews, we also extract various kinds of 
relevant metadata, made available in a JSON representation with normalized 
attribute--value names across reviews. 
Metadata in NoReC include information like the URL of the originally published
document, numerical rating, publishing date, author list, domain or
thematic category, original ID in the source, and more. Beyond this we also add
information about the identified language variety (Bokmål/Nynorsk), the 
assigned data split (test/dev/train, as further described in 
Section~\ref{sec:distribution}), the assigned document ID, and finally a
normalized thematic category. 

\paragraph{Thematic categories} 
\begin{table}
\centering
\begin{smaller}
\begin{tabular}{@{}lr@{}}
\toprule
Category & \# Reviews \\
\midrule
Screen & 13,085 \\
Music & 12,410 \\
Literature & 3,530 \\
Products & 3,120 \\
Games & 1,765 \\
Restaurants & 534 \\
Stage & 530 \\
Sports & 118 \\
Misc & 102 \\
\midrule
Total & 35,194 \\
\bottomrule
\end{tabular}
\caption{Number of reviews across categories.}
\label{tab:categories}
\end{smaller}
\end{table}

The `category' attribute warrants some elaboration.  The use of thematic
categories and/or tags varies a lot between the different sources, ranging from
highly granular categories to umbrella categories encompassing many different
domains. Based on the original inventory of categories, each review in NoReC is
mapped to one out of nine normalized thematic categories with English names.
The distribution over categories is shown in Table~\ref{tab:categories}, sorted
by frequency.

For some sources, this normalization is a matter of simple one-to-one mapping,
while for others it is more complex, involving heuristics based on the
presence of certain tags and keywords in the title. The granularity in the
final set of categories is limited by the granularity in the sources. However,
the original (Norwegian) source categories are also preserved in the metadata 
(`source-category'). 

As seen from Table~\ref{tab:categories}, the two categories that are by far the
largest are `screen' and `music'. While the former covers reviews about movies
and TV-series, 
the latter covers both musical recordings and performances. The related
category `stage' covers theater, opera, ballet, musical and other stage
performances besides music. The perhaps most diverse category is `products',
which comprises product reviews across a number of sub-categories, ranging from
cars and boats to mobile phones and home electronics, in addition to travel and
more. The remaining categories of `literature', `games', `restaurants', and
`sports' are self-explanatory, while the `misc' category was included to cover
topics that were infrequent or that could not easily be mapped to any of the
other categories by simple heuristics.  

\paragraph{Ratings} 

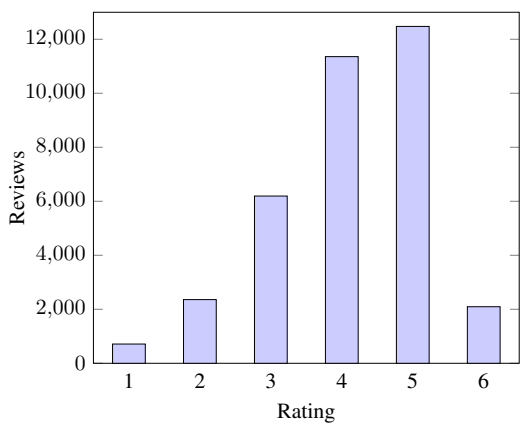
\begin{figure}[t!]
\begin{center}
\resizebox{0.4\textwidth}{!}{%
  \begin{tikzpicture}
\begin{axis}[
  xlabel=Rating,
  ylabel=Reviews,
  ymin=0,
  ymax=13000,
  scaled y ticks = false,
  symbolic x coords={1, 2, 3, 4, 5, 6},
  xtick=data,
  xtick style={draw=none},
  xticklabel style={anchor=base,yshift=-7pt},
  ybar, 
  bar width=15pt,
  ]
  \addplot[
    fill = blue!20
  ]
  coordinates{
    (1,713)(2,2359)(3,6194)(4,11357)(5,12476)(6,2095)
  };
  \end{axis}
\end{tikzpicture}
}\vspace{-2ex}
\caption{Number of reviews across ratings.}
\label{tab:ratings}
\end{center}
\end{figure}

From the perspective of SA, the most immediately relevant piece of metadata is
obviously the rating. As discussed previously, all the reviews were originally
published with an integer-valued rating between 1 and 6, visually indicated
using the face of a die. 
Figure~\ref{tab:ratings} shows the
distribution of reviews relative to rating scores. We see that rating values of
4 and 5 are the most common, while rather few reviews were given the lowest
possible rating of 1.  
For the final version of the paper we plan to report more details on the rating
distributions, for example investigating whether there are differences in the
use of the rating scale between different sources, domains, or authors. For
future work it would also be interesting to try to uncover whether any rating
biases exists, for example related to gender.

\subsection{Formats and availability}
\label{sec:distribution}

Distributed under a CC BY-NC 4.0 
license,\footnote{\url{https://creativecommons.org/licenses/by-nc/4.0/}} NoReC
is available for download from the following git repository:\\[2pt]
\url{https://github.com/ltgoslo/norec}
\paragraph{Formats} NoReC distributes two formats. The first is the \conllu\
format as described in Section~\ref{sec:enrichments}, containing sentence
segmented and tokenized text annotated with PoS tags and dependency
graphs. This is considered the primary format. Secondly, we also distribute the
canonical HTML representation of the `raw' review documents as described in
Section~\ref{sec:content-to-html}.  For each format, each review is stored as a
separate file, with the filename given by the review ID.  To facilitate a low
barrier of use for different types of end-users, we also include scripts for
converting from \conllu\ to running tokenized text (using either full-forms or
lemmas) and from HTML to raw text without pre-processing. The metadata for each
review is provided as a JSON object, all listed in a single file and indexed on
the document IDs. 
The NoReC git repository will include a Python module with basic functionality
for reading the \conllu\ and JSON representations, as to make experimentation
with the corpus as accessible and convenient as possible. 

% \paragraph{Train/dev/test splits} To facilitate replicability of experiments
% the corpus comes with pre-defined standard splits for training, development and
% testing, with a 80--10--10 ratio. The splits were created by reserving every
% 9th review for the development set set and every 10th for the test set, with
% the rest being assigned to the training set. This simple scheme ensures a 
% balanced selection across all relevant variables; domain, source, rating,
% author, and so on. 

\paragraph{Train/dev/test splits} To facilitate replicability of experiments,
NoReC comes with pre-defined splits for training, development and
testing. These were defined by first sorting all reviews for each category by
publishing date and then reserving the first 80\% for training, the subsequent
10\% for development, and the final 10\% for held-out testing. This splitting
strategy makes the test setting as realistic as possible and avoids having
multiple reviews for the same product (from different sources) across splits.

\section{Summary and outlook} 
\label{sec:summary}

The current paper describes the creation of the Norwegian Review Corpus;
NoReC. The final dataset comprises more than 35,000 full-text reviews 
($\approx$ 15 million tokens) from a  
wide range of different domains, collected from several major Norwegian 
news sources. Each review is rated with a numerical score on a scale of 1--6,
and can be used for training and evaluating models for document-level sentiment
analysis. 
While the primary distribution format of the corpus is \conllu\ -- based
on only the extracted text and applying UDPipe for a full pre-processing 
pipeline from sentence segmentation to dependency parsing -- the release also includes HTML 
representations of the full reviews with all content preserved. 
Each review is in addition associated with a rich set of metadata, 
including thematic category, numerical rating, and more. We also provide 
pre-defined splits for training, development and testing. 

For future work, the SANT project will seek to build on NoReC to (i) experiment
with both polarity classification and rating inference on the document-level
using neural architectures, (ii) extract SA lexicons encoding the polarity of
individual words, and finally (iii) also move beyond the document-level and
manually add more fine-grained and aspect-based SA annotations for a sub-set of
the corpus.  Across all these activities, the various thematic categories will
be useful for assessing cross-domain effects (e.g., how well does an SA
classifier trained on movie reviews perform for home electronics?) and
potentially even for training domain-specific models. It will of course also be
important to asses how well SA resources developed on the basis of the reviews
generalize to non-review texts.

Resources for sentiment analysis have so far been unavailable for Norwegian. As
such, NoReC represents a highly sought-after addition to Norwegian language
technology, valuable to both industry and the research community alike.

\section{Acknowledgments}
This work was carried out as part of the SANT research project (Sentiment
Analysis for Norwegian Text), funded by an IKTPLUSS grant from the Research
Council of Norway (project no. 270908). SANT represents a newly initiated
collaboration between the Department of Informatics at the University of Oslo,
and three of Norway's largest media groups; the Norwegian Broadcasting
Corporation, Schibsted Media Group and Aller Media.

\section{Bibliographical references}
\bibliographystyle{lrec}
\bibliography{ltg.bib}

\providecommand{\fromto}[2]{#1$\,$--$\,$#2}\providecommand\Beek[1]{\mbox{#1}}\providecommand\Noord[1]{\mbox{#1}}\providecommand\Lohuizen[1]{\mbox{#1}}
\begin{thebibliography}{}

\bibitem[\protect\citename{Blitzer \bgroup et al.\egroup }2007]{Bli:Dre:Per:07}
Blitzer, J., Dredze, M., and Pereira, F.
\newblock (2007).
\newblock Biographies, bollywood, boom-boxes and blenders: Domain adaptation
  for sentiment classification.
\newblock In {\em Proceedings of the 45th {M}eeting of the {A}ssociation for
  {C}omputational {L}inguistics}, page \fromto{187}{205}, Prague, Czech
  Republic.

\bibitem[\protect\citename{Lui and Baldwin}2012]{Lui:Bal:12}
Lui, M. and Baldwin, T.
\newblock (2012).
\newblock {langid.py}: An off-the-shelf language identification tool.
\newblock In {\em Proceedings of the 50th {M}eeting of the {A}ssociation for
  {C}omputational {L}inguistics {S}ystem {D}emonstrations}, page
  \fromto{25}{30}, Jeju, Republic of Korea.

\bibitem[\protect\citename{Maas \bgroup et al.\egroup }2011]{Maa:Dal:Pha:11}
Maas, A.~L., Daly, R.~E., Pham, P.~T., Huang, D., Ng, A.~Y., and Potts, C.
\newblock (2011).
\newblock Learning word vectors for sentiment analysis.
\newblock In {\em Proceedings of the 49th {M}eeting of the {A}ssociation for
  {C}omputational {L}inguistics}, page \fromto{142}{150}, Portland, OR, USA.

\bibitem[\protect\citename{Nivre \bgroup et al.\egroup }2016]{Niv:Mar:Gin:16}
Nivre, J., de~Marneffe, M.-C., Ginter, F., Goldberg, Y., Haji\v{c}, J.,
  Manning, C.~D., McDonald, R., Petrov, S., Pyysalo, S., Silveira, N.,
  Tsarfaty, R., and Zeman, D.
\newblock (2016).
\newblock Universal dependencies v1: A multilingual treebank collection.
\newblock In {\em Proceedings of the 10th {I}nternational {C}onference on
  {L}anguage {R}esources and {E}valuation}, Portoro\v{z}, Slovenia.

\bibitem[\protect\citename{{\O}vrelid and Hohle}2016]{Ovr:Hoh:16}
{\O}vrelid, L. and Hohle, P.
\newblock (2016).
\newblock {N}orwegian {U}niversal {D}ependencies.
\newblock In {\em Proceedings of the 10th {I}nternational {C}onference on
  {L}anguage {R}esources and {E}valuation}, page \fromto{1579}{1585},
  Portoro\v{z}, Slovenia.

\bibitem[\protect\citename{Pang and Lee}2005]{Pan:Lee:05}
Pang, B. and Lee, L.
\newblock (2005).
\newblock Seeing stars: Exploiting class relationships for sentiment
  categorization with respect to rating scales.
\newblock In {\em Proceedings of the 43rd {M}eeting of the {A}ssociation for
  {C}omputational {L}inguistics}, page \fromto{115}{124}, Ann Arbor, MI, USA.

\bibitem[\protect\citename{Socher \bgroup et al.\egroup }2013]{Soc:Per:Wu:13}
Socher, R., Perelygin, A., Wu, J., Chuang, J., Manning, C., Ng, A., and Potts,
  C.
\newblock (2013).
\newblock Recursive deep models for semantic compositionality over a sentiment
  treebank.
\newblock In {\em Proceedings of the 2013 {C}onference on {E}mpirical {M}ethods
  in {N}atural {L}anguage {P}rocessing}, page \fromto{1631}{1642}, Seattle, WA,
  USA.

\bibitem[\protect\citename{Straka and Strakov\'{a}}2017]{Str:Str:17}
Straka, M. and Strakov\'{a}, J.
\newblock (2017).
\newblock Tokenizing, {POS} tagging, lemmatizing and parsing {UD} 2.0 with
  {UDPipe}.
\newblock In {\em Proceedings of the CoNLL 2017 Shared Task: Multilingual
  Parsing from Raw Text to Universal Dependencies}, pages 88--99, Vancouver,
  Canada.

\bibitem[\protect\citename{Straka \bgroup et al.\egroup }2016]{Str:Haj:Str:16}
Straka, M., Haji\v{c}, J., and Strakov\'{a}, J.
\newblock (2016).
\newblock {UDPipe:} trainable pipeline for processing {CoNLL-U} files
  performing tokenization, morphological analysis, pos tagging and parsing.
\newblock In {\em Proceedings of the Tenth International Conference on Language
  Resources and Evaluation}, Portoro\v{z}, Slovenia.

\bibitem[\protect\citename{Velldal \bgroup et al.\egroup }2017]{Vel:Ovr:Hoh:17}
Velldal, E., Øvrelid, L., and Hohle, P.
\newblock (2017).
\newblock Joint {UD} parsing of {N}orwegian {B}okmål and {N}ynorsk.
\newblock In {\em Proceedings of the 11th {N}ordic {C}onference of
  {C}omputational {L}inguistics}, page \fromto{1}{10}, Gothenburg, Sweden.

\bibitem[\protect\citename{Wang \bgroup et al.\egroup }2010]{Wan:Lu:Zha:10}
Wang, H., Lu, Y., and Zhai, C.
\newblock (2010).
\newblock Latent aspect rating analysis on review text data: A rating
  regression approach.
\newblock In {\em Proceedings of the 16th ACM SIGKDD Conference on Knowledge
  Discovery and Data Mining (KDD'2010)}, page \fromto{783}{792}.

\bibitem[\protect\citename{Zeman \bgroup et al.\egroup }2017]{Zem:Pop:Str:17}
Zeman, D., Popel, M., Straka, M., Hajic, J., Nivre, J., Ginter, F., Luotolahti,
  J., Pyysalo, S., Petrov, S., Potthast, M., Tyers, F., Badmaeva, E., Gokirmak,
  M., Nedoluzhko, A., Cinkova, S., Hajic~jr., J., Hlavacova, J.,
  Kettnerov\'{a}, V., Uresova, Z., Kanerva, J., Ojala, S., Missil\"{a}, A.,
  Manning, C.~D., Schuster, S., Reddy, S., Taji, D., Habash, N., Leung, H.,
  de~Marneffe, M.-C., Sanguinetti, M., Simi, M., Kanayama, H., dePaiva, V.,
  Droganova, K., Mart\'{i}nez~Alonso, H., \c{C}\"{o}ltekin, c., Sulubacak, U.,
  Uszkoreit, H., Macketanz, V., Burchardt, A., Harris, K., Marheinecke, K.,
  Rehm, G., Kayadelen, T., Attia, M., Elkahky, A., Yu, Z., Pitler, E.,
  Lertpradit, S., Mandl, M., Kirchner, J., Alcalde, H.~F., Strnadov\'{a}, J.,
  Banerjee, E., Manurung, R., Stella, A., Shimada, A., Kwak, S., Mendonca, G.,
  Lando, T., Nitisaroj, R., and Li, J.
\newblock (2017).
\newblock Conll 2017 shared task: Multilingual parsing from raw text to
  universal dependencies.
\newblock In {\em Proceedings of the {CoNLL} 2017 Shared Task: Multilingual
  Parsing from Raw Text to {Universal} {Dependencies}}, page \fromto{1}{19},
  Vancouver, Canada.

\end{thebibliography}

\end{document}